\documentclass[letterpaper, 10 pt, conference]{ieeeconf}  
\usepackage[letterpaper, left=1in, right=0.5in, bottom=0.75in, top=0.8in]{geometry}
                                                          
\IEEEoverridecommandlockouts                              
                                               
\usepackage{graphicx}
\usepackage{amsmath}
\usepackage{bm}
\usepackage{url}
\usepackage{subcaption}
\usepackage{tabularx}
\usepackage[normalem]{ulem}
\usepackage{xcolor}

\title{\LARGE \bf
Investigating Mixed Reality for Communication Between Humans and Mobile Manipulators}

\author{Mohamad Shaaban, Simone~Macci\`o, Alessandro~Carf\`i, and Fulvio~Mastrogiovanni
\thanks{All authors are with TheEngineRoom lab, part of the Department of Informatics, Bioengineering, Robotics, and Systems Engineering (DIBRIS), University of Genoa, Via Opera Pia 13, 16145 Genoa, Italy. Corresponding author's email: mohamad.shaaban@edu.unige.it}%
}

\begin{document}

\maketitle

\begin{abstract}

This article investigates mixed reality (MR) to enhance human-robot collaboration (HRC). The proposed solution adopts MR as a communication layer to convey a mobile manipulator's intentions and upcoming actions to the humans with whom it interacts, thus improving their collaboration. A user study involving 20 participants demonstrated the effectiveness of this MR-focused approach in facilitating collaborative tasks, with a positive effect on overall collaboration performances and human satisfaction. 
\end{abstract}

\begin{keywords}
Human-Robot Collaboration, Mixed Reality, Software Architecture.
\end{keywords}

\section{Introduction}
\label{section1}

Human-robot collaboration (HRC) is becoming increasingly important in the era of human-centred production and smart factories, both from scientific and industrial standpoints. 
In HRC, humans and robots share physical space and duties \cite{arents2021human, Muralietal2020}, combining the benefits of human cognitive ability with machine efficiency and speed. 

However, this new paradigm introduces several technical challenges,  from ensuring human teammate's safety throughout the collaboration process to developing efficient communication interfaces for hybrid human-robot teams. Efforts to enhance human safety in human-robot collaboration (HRC) have led to various solutions. These include strategies to minimize collision risks by predicting human space occupancy \cite{ragaglia2018trajectory}, as well as techniques for detecting and mitigating contact \cite{roveda2020model, ren2018collision}. Communication also plays a vital role in ensuring smooth interactions. Sharing information about the robot's intents is important and enables humans to anticipate and react accordingly. Researchers have investigated several methods to provide humans with feedback about the robot's internal state either through visual \cite{wengefeld2020laser} or vocal \cite{nikolaidis2018planning} feedback during HRC processes.
 
\begin{figure}[t!]
\centering
\begin{subfigure}{.48\textwidth}
\centering
\includegraphics[width=\linewidth]{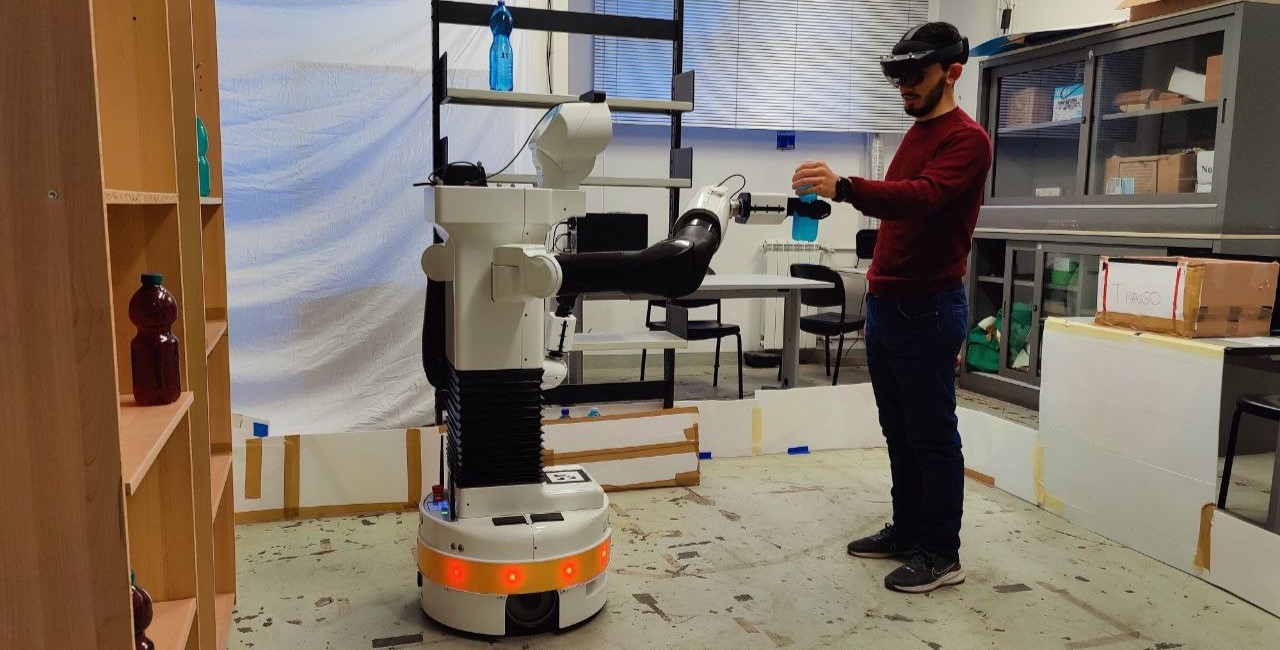}
\caption{}
\label{fig:hrc}
\end{subfigure}
\begin{subfigure}{.48\textwidth}
\centering
\includegraphics[width=\linewidth]{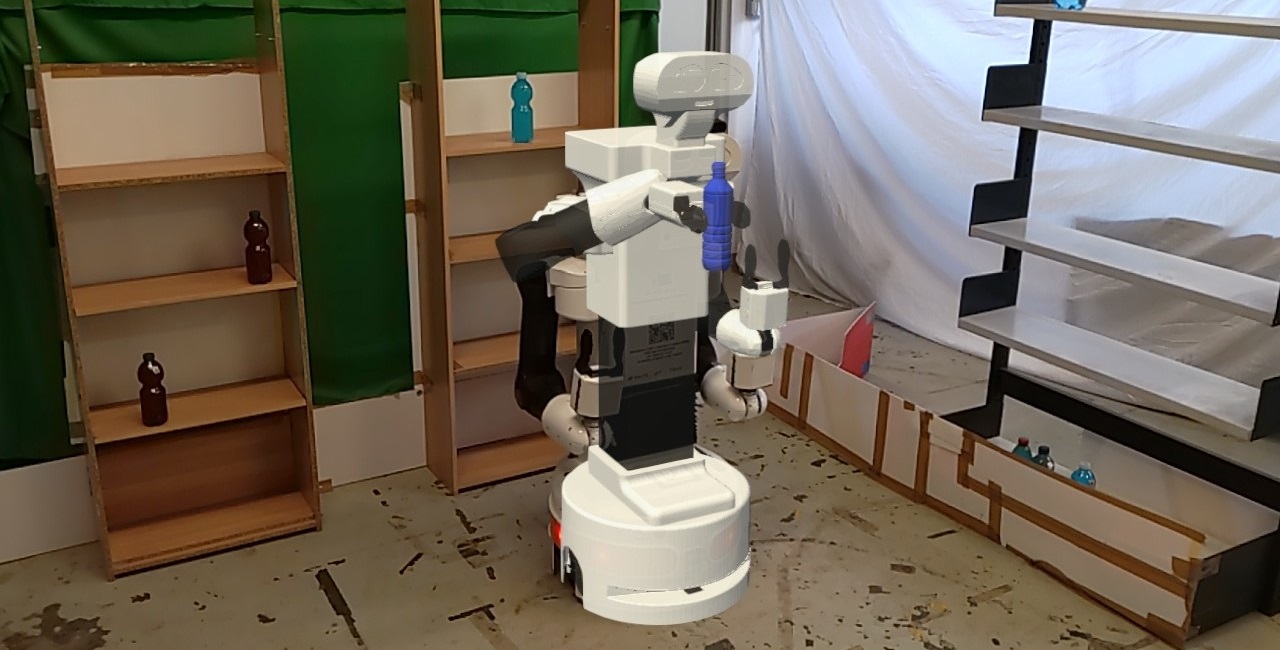} 
\caption{}
\label{fig:dt}
\end{subfigure}
\caption{
{Fig. \ref{fig:hrc} illustrates the scenario of human-robot collaboration, setting the stage for understanding the context. Following this, Fig. \ref{fig:dt} offers a view from the human perspective within mixed reality. This perspective demonstrates how the handover process will occur by conveying the robot's intention, thereby ensuring human safety.}}
\label{fig:hrc-dt}
\end{figure}

In this work, we explore how Mixed Reality (MR), as a communication medium that projects holographic representations of data into the real environment in a coherent, contextual way, can enhance human-robot collaboration (HRC). MR has been explored as a tool for effective communication between humans and robots, particularly using head-mounted displays (HMDs). These devices hold the potential to make the robot's internal state intuitively understandable to human collaborators by projecting its planned actions as holograms. By understanding a robot's intended actions, humans can better coordinate their own, leading to smoother and safer interactions. This clarity reduces the risk of accidents and ensures both parties can work more efficiently together.

\begin{figure*}[t!]
\centering 
\includegraphics[width=0.8\textwidth]{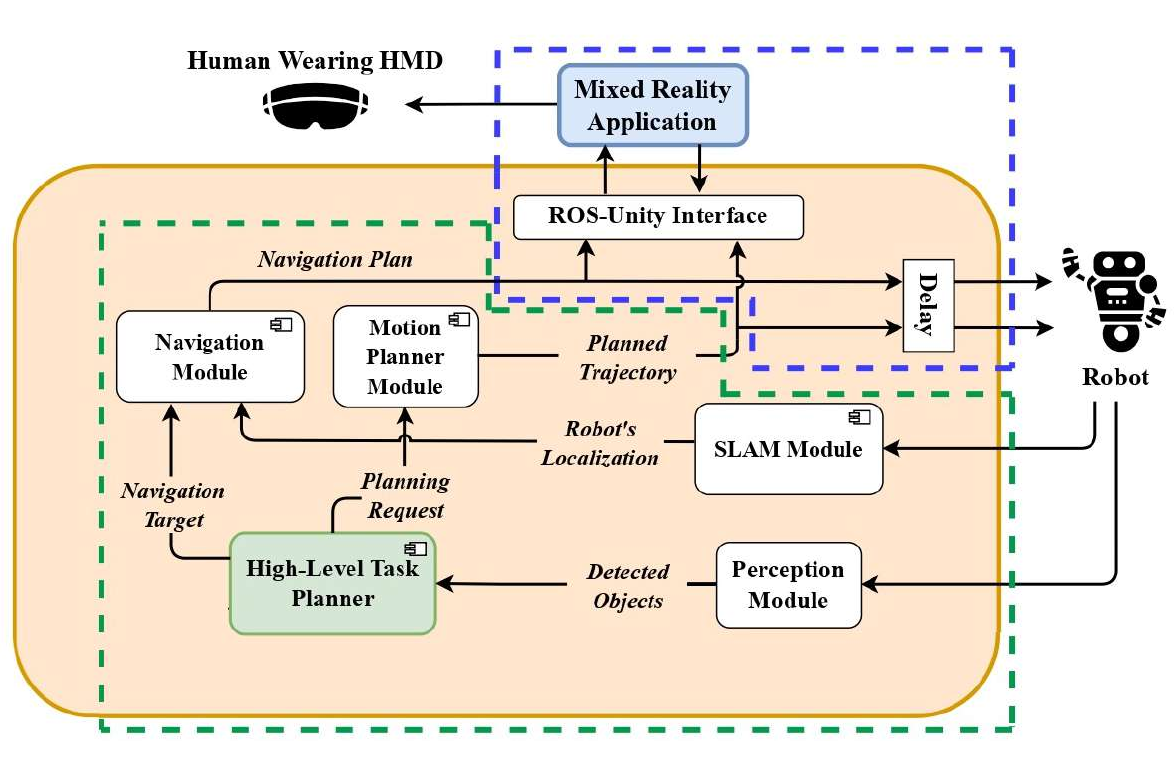}
\caption{
A detailed overview of MR-HRC v2, with the two main blocks highlighted by the corresponding colours.}
\label{fig:systemarchitecturefigure}
\end{figure*}

In a previous article, we introduced the MR-HRC architecture,  which leverages the MR layer to display a robot manipulator's \textit{intentions}, defined as upcoming planned actions, through holographic cues \cite{maccio2022mixed}. With a user study in a collaborative scenario, we proved how such a form of holographic communication benefited the overall performance of the human-robot team. Nevertheless, the scheme proposed in \cite{maccio2022mixed} was limited to only rendering intentions for fixed manipulator robots. On the contrary, the present work aims to overcome these limitations extending the developed framework through an expanded architecture named MR-HRC v2, which permits the communication of mobile manipulators intent over the MR Layer. Although our work focuses on the communication aspect brought by MR, we also observe a contribution to the safety problem, derived from the additional safety induced by an intuitive, holographic communication of robot motions throughout an HRC process.

To validate the effectiveness of the expanded architecture, we carried out an experimental campaign with 20 participants in a complex scenario of mobile collaboration, where a human and a robot were required to interact while also carrying out independent, concurrent activities in a scenario simulating a logistic centre. The user study proved the architecture's effectiveness in driving collaboration. Furthermore, we observed a positive effect of MR-induced communication on task quality and team efficiency.


\begin{figure*}[t!]
\centering 
\begin{subfigure}{.4\textwidth}
  \centering
  \includegraphics[width=\linewidth]{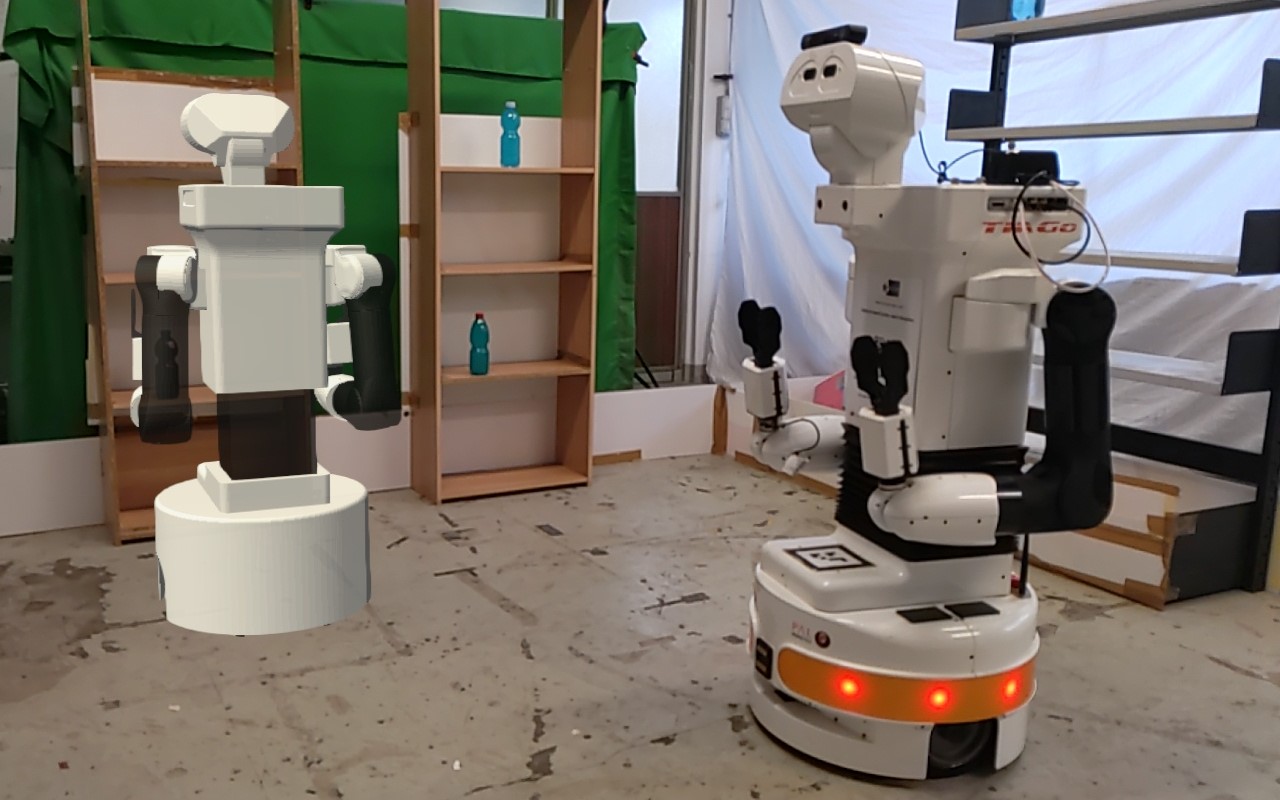}  
  \caption{}
  \label{holo1}
\end{subfigure}
\begin{subfigure}{.4\textwidth}
  \centering
  \includegraphics[width=\linewidth]{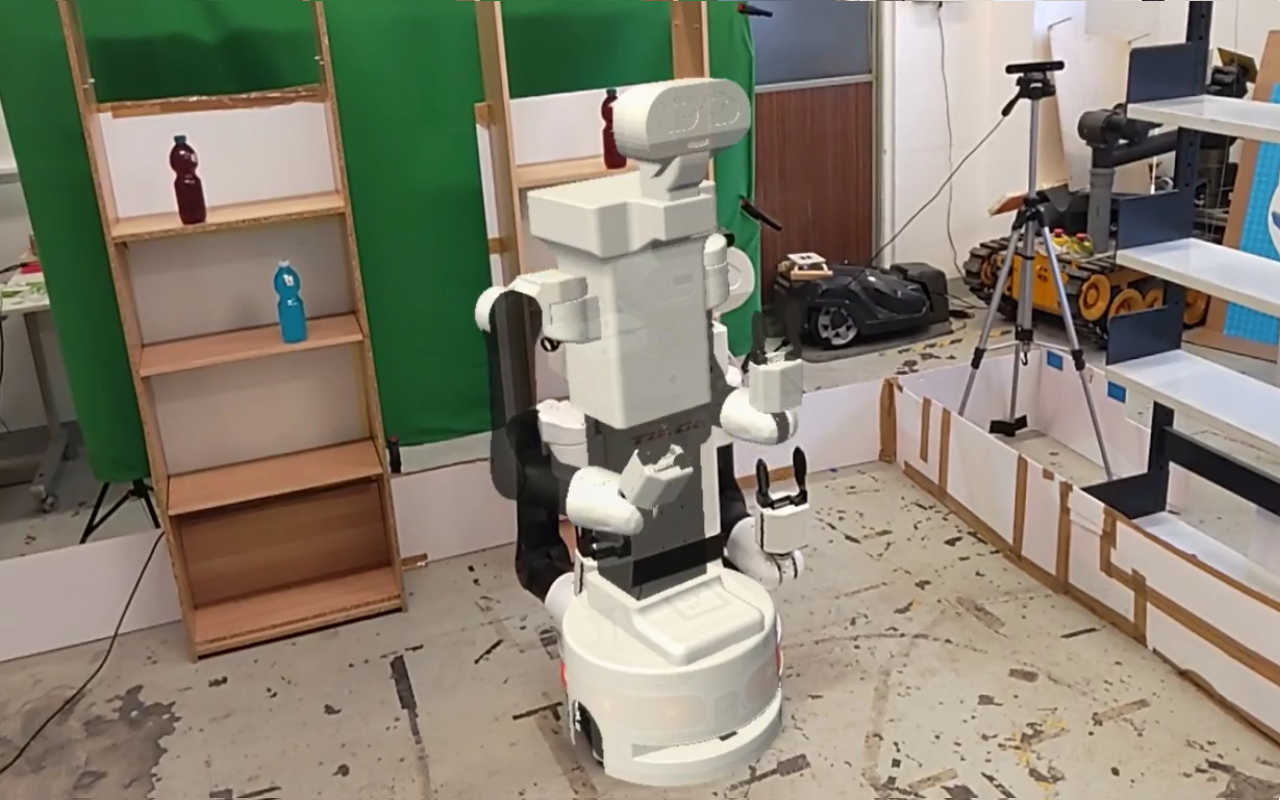}  
  \caption{}
  \label{holo2}
\end{subfigure}
\begin{subfigure}{.4\textwidth}
  \centering
  \includegraphics[width=\linewidth]{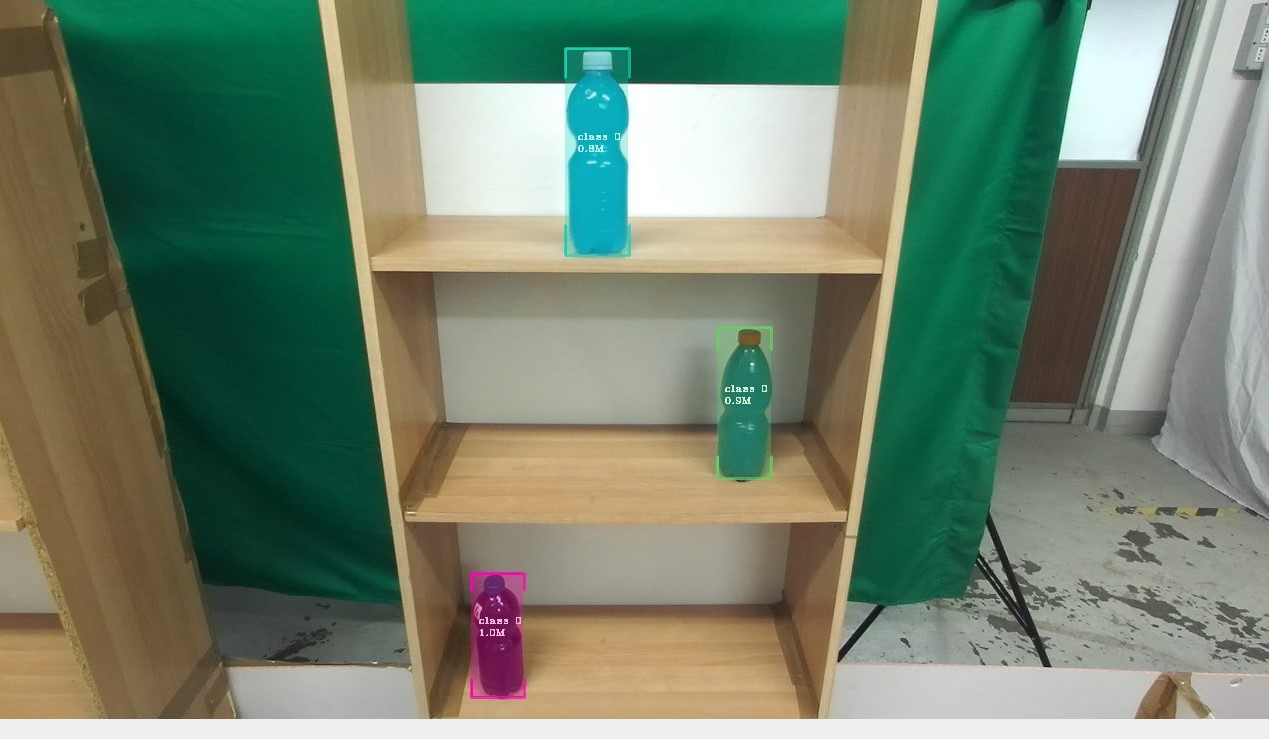}  
  \caption{}
  \label{holo3}
\end{subfigure}
\begin{subfigure}{.4\textwidth}
  \centering
  \includegraphics[width=\linewidth]{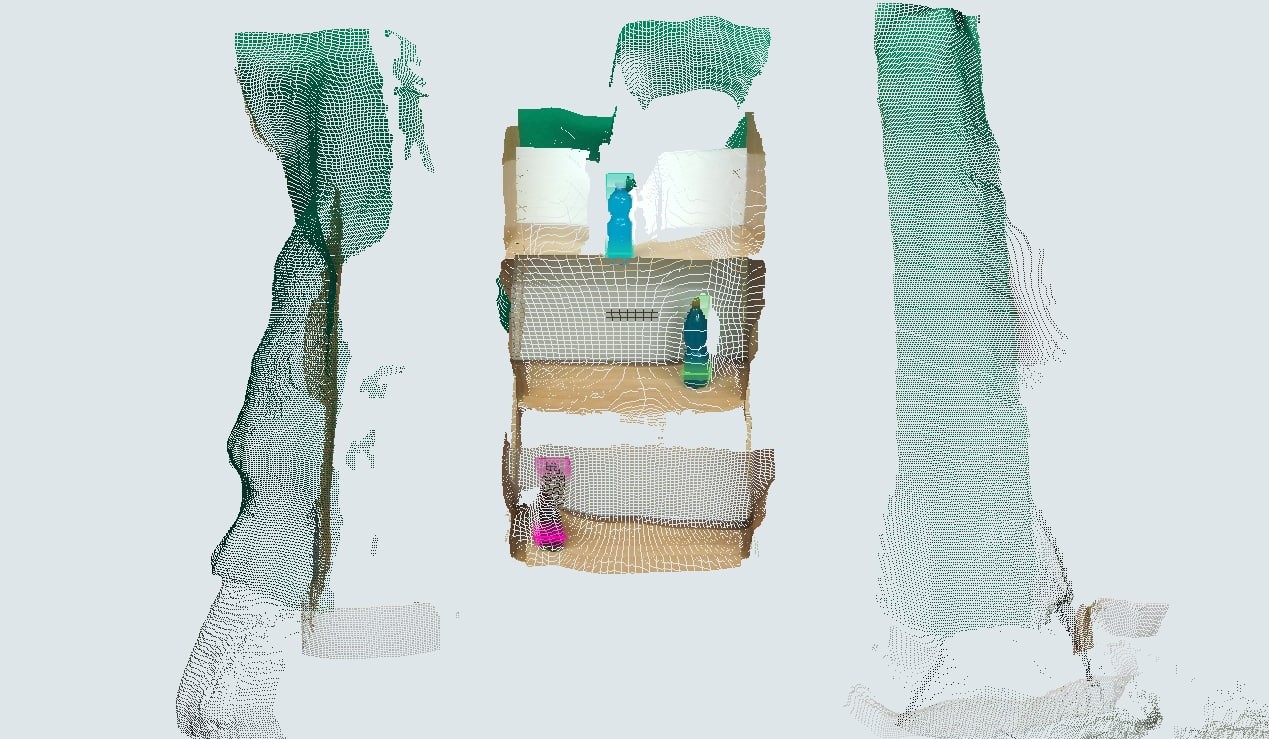}  
  \caption{}
  \label{ext1}
\end{subfigure}
\caption{
Fig. \ref{holo1} and \ref{holo2} show the holographic communication from the human perspective, respectively, for robot navigation and manipulation. 
Fig. \ref{holo3} and \ref{ext1} present the results of the perception module, with the bottles detected by YOLOv5m and the corresponding point cloud acquired by the ZED2 camera mounted on the robot.}
\label{fig:exp_screenshots}
\end{figure*}

\section{Related Work}
\label{relatedwork}

Effective communication between humans and robots is crucial for fluent collaboration, enabling agents to coordinate, synchronize, and achieve seamless interactions. HRC can require explicit and implicit communication and leverage visual, auditory, and tactile channels. 
Commonly adopted solutions are based on gestural interaction, speech, and haptic feedback \cite{pascher2023communicate, skantze2021turn}. 
Verbal communication offers an intuitive and expressive solution to convey robots' intentions to the operator \cite{nikolaidis2018planning}. However, it can be easily affected by background noise, which is frequent in industrial settings. 
Various approaches using visual stimuli have been proposed in the literature. 
Blinking lights and flashing cues \cite{hetherington2021hey}, or screens located in the workspace, may require high levels of effort to be fully understood by humans.
On the contrary, 2D projections on relevant objects are more intuitive but require structured environments \cite{costa2022augmented}. 
Augmented reality handheld devices enable the overlay of holograms to the environment, allowing an additional communication modality. However, since operators must hold the device, it compromises their ability to work freely \cite{chacko2019augmented}. 
Conversely, by exploiting MR-based HMD devices, researchers could project static and dynamic holograms in the user's first-person view while ensuring the operator's hands are free \cite{maccio2022mixed, calzone2022mixed}.

The effectiveness of HMD to display in advance the robot's planned motion has been demonstrated in studies of collaborative interactions between humans and a dual-arm manipulation system, such as Baxter from Rethink Robotics \cite{rosen2019communicating}. Greenfield \textit{et al}. (2020) extended the research on intent communication through mixed reality by introducing volume visualization alongside path visualization to indicate the space the robot will occupy. They used colour gradients to represent proximity, allowing users to understand safe zones \cite{gruenefeld2020mind}. They validated this solution through an experimental study using a KUKA LBR iiwa 7 R800 manipulator in a collaborative setting. The results showed higher perceived safety among participants, enabling them to remain close to the robot without altering its operations.
Additionally, in a more recent study, Newbury \textit{et al}., 2021 an HMD to convey the robot's intent and perception pipeline output for synchronizing human-robot interactions during handovers \cite{newbury2021visualizing}. Notably, the study introduced an innovative method utilizing holographic overlays to display the estimated object pose and intended grasp pose of a Franka Emika Panda manipulator, demonstrating its effectiveness in enhancing interaction safety, trustworthiness, fluency, and predictability.

These studies have significantly advanced the field of intent communication in HRC using mixed reality. However, their primary focus was fixed manipulators such as Rethink Robotics Baxter and KUKA LBR iiwa 7 R800. These studies showcase the potential of MR and AR technologies in enhancing safety, efficiency, and understanding between humans and robots in a shared workspace. However, mobile robots introduce different challenges and opportunities for intent communication, such as dynamic environment navigation and spatial awareness. Although HMD received significant attention regarding mobile robots' teleoperation and programming \cite{walker2023virtual}, really few works address the intent communication aspect. While, Walker \textit{et al}., 2018 proposed an MR framework designed to communicate drone motion intent, improving user understanding of robot goals 
 \cite{walker2018communicating}, Zu \textit{et al}., 2018 presented an MR interface that visualizes a robot sensory data (e.g., laser scanner and cost map) alongside static planned paths and handover positions \cite{zu2018improving}. This article seeks to contribute to this research by extending an existing architecture \cite{maccio2022mixed} to suit the mobile manipulator's scenario by dynamically displaying both navigation and manipulation intentions. This approach enhances the human ability to supervise robots' activities and improves safety by visualizing the future robot's state. Additionally, we conduct a preliminary experimental study to assess the effectiveness of MR intent communication for mobile manipulators.


\section{System's Architecture}
\label{sec:softarchitecture}

The MR-HRC v2 architecture, shown in Fig. \ref{fig:systemarchitecturefigure}, comprises two primary components: the Mixed Reality (MR) Application, highlighted in blue, and the robot application core, represented in green. This setup facilitates interaction between a human operator and a mobile manipulator by using an MR headset to convey the robot’s intentions to the human, focusing on a direct, intuitive communication method to represent the state of collaboration and the robot's actions.

The part of the architecture handling robot operations follows the classical sense-reason-act paradigm. 
In this context, localization and recognition of objects (see Fig. \ref{holo3}) and robot self-localization in the environment are handled online by the \textit{Perception} and \textit{SLAM} modules using onboard sensors.
Then the \textit{High-Level Task Planner} module plans the sequence of actions (e.g., \textit{pick a bottle} or \textit{navigate to shelf}) based on the perceived objects, and the final goal. 
Actions resulting from the \textit{High-Level Task Planner} are handled by the \textit{Motion Planner} when they involve manipulations or by the \textit{Navigation Planner} when they are navigation tasks.

The \textit{Mixed Reality Application} communicates to the human the intentions of the robot using dynamic holograms that depict its future state (see Fig. \ref{holo1} and \ref{holo2}). 
This approach was first introduced by Macciò \textit{et al}., 2022 \cite{maccio2022mixed}, and it is made possible by applying a temporal delay ($\Delta t$) to the robot commands generated by the \textit{Navigation Planner} and the \textit{Motion Planner}. 


\section{Implementation, Frameworks, and Equipment}
\label{sec:implementation}

We implemented the MR-HRC v2 architecture, presented in the previous section, by embracing the open-source paradigm and making tailored contributions to relevant projects as needed. We chose the Robot Operating System (ROS) framework as our implementation platform and, for validation purposes, integrated with TIAGo++, a robot from Pal Robotics. Here, we describe the implementation details for each module in the architecture, highlighting our contributions.

Robot operations are controlled with already available software modules. 
The robot's arms motion planning is performed via \textit{Moveit}, whereas autonomous localization and navigation are handled by the \textit{ROS Navigation Stack}. 
We extended TIAGo's original perception capabilities to detect and localize objects in the environment. 
On top of the robot's head, we mounted a ZED2 camera, calibrated with respect to the robot's reference frame and providing a wide-angle field of view (FOV). 
Whenever a new frame from the ZED2's left camera is captured, it is processed with YOLOv5 to recognize and localize objects in 2D space. 
The 3D pose of an object is estimated by projecting its 2D bounding box on the corresponding depth map, see Fig. \ref{ext1}. 
The result is an array of 3D bounding boxes 
that are used for planning pick actions.
The \textit{Perception Module} runs on an NVIDIA JETSON TX2, with 256 CUDA cores, mounted on the robot. 
At this stage, the \textit{High-Level Task Planner} holds a predefined sequential set of actions for the robot to perform, i.e., \textit{pick}, \textit{transport}, and \textit{place} actions. 
At run time, the predefined actions are grounded with values from the perception system, i.e., the object positions.
The high-level task planner also describes how the human and the robot interact. In particular, when an object is out of the robot's reachable workspace, the \textit{High-Level Task Planner} activates the "handover mode". In this process,  the robot executes a half-turn rotation around its base before assuming a predetermined handover pose (see Fig. \ref{fig:hrc-dt}). It then extends its right manipulator, opens its gripper, and waits for the human to hand over the desired object.

The \textit{Mixed Reality Application}, developed using Unity and deployed to Hololens 2, a Microsoft state-of-the-art MR-HMD device, runs at 30Hz with the native Hololens resolution.
On top of the 3D engine, we use two SDKs, namely \textit{Vuforia}, which is currently used to extract the robot position in the MR device reference frame using a 25h9 April tag attached to the robot base, and Microsoft \textit{Mixed Reality Toolkit} (MRTK), responsible for overlaying 3D holograms on top of the real world. 
In this implementation of the MR communication, the delays $\Delta t$ between the hologram and the robot motions have been set to 5 seconds,
a value empirically determined to allow a reasonable separation between holographic and subsequent robot movements without significantly affecting task pace.
Unity official support is provided for interfacing with ROS\footnote{\url{github.com/Unity-Technologies/ROS-TCP-Endpoint}}, thus providing proper communication with the rest of the architecture. 
The MR application's source code is openly available to other researchers through GitHub\footnote{\url{github.com/TheEngineRoom-UniGe/MR-Tiago}}. 


\section{Experimental Setup} 
\label{sec:experimentalsetup}
\begin{figure}[t!]
\centering
\includegraphics[width=0.95\columnwidth]{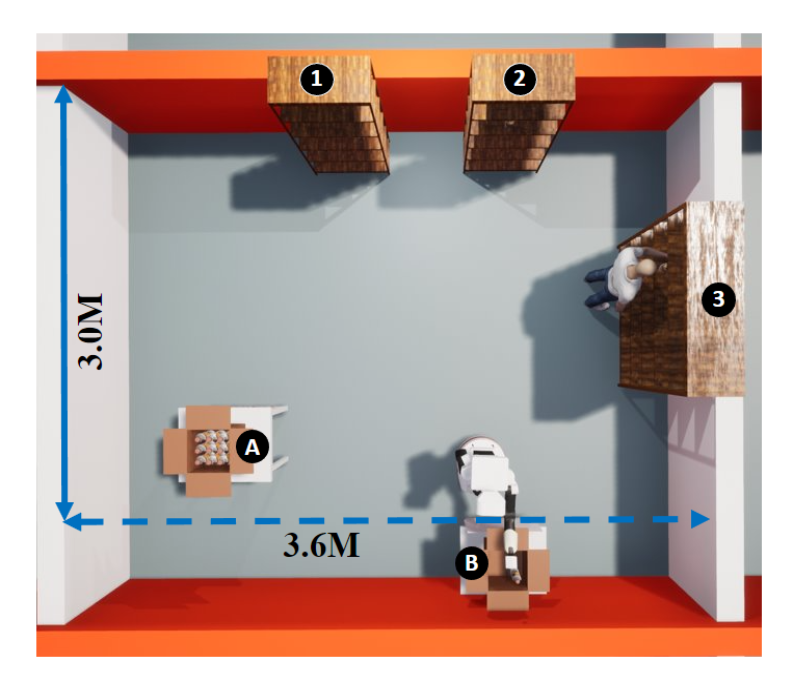}
\caption{
Bird's-eye view of the collaborative workspace, displaying labelled points of interest alongside the workspace dimensions.
}
\label{fig::experiment}
\end{figure}
The experiments detailed in this section aim to evaluate the impact of using Mixed Reality (MR) technology for communicating robot intentions in human-robot collaboration (HRC). Previous research has demonstrated the positive influence on HRC when MR is used to convey the intentions of static manipulators \cite{maccio2022mixed}  and drones \cite{walker2018communicating} . Building on this foundation, our study extends the investigation to a more complex scenario involving mobile manipulators.
Therefore, in the context of this work, we hypothesized that the adoption of MR technologies in HRC with a mobile manipulator could lead to:

\begin{itemize}
\item[\textit{H1}] Reduced overall task completion time - We expect MR to streamline the human-robot interaction, leading to faster competition times.
\item[\textit{H2}] Enhanced team fluency - MR can foster smoother collaboration by:
\begin{itemize}
\item[\textit{H2.1}] Increasing proactive human interventions to assist the robot - By providing real-time information and guidance through MR, humans can anticipate the robot's needs and offer assistance proactively.
\item[\textit{H2.2}] Decreasing the number of failed interactions - Clearer communication and shared situational awareness through MR should minimize misunderstandings and lead to fewer failed interactions.
\item[\textit{H2.3}] Decreasing the times in which human operator should pause their task to observe the robot - MR can keep humans informed about the robot's progress without requiring them to constantly monitor its actions.
\end{itemize}
\end{itemize}
\color{black}

\subsection{Collaborative Scenario} 
\label{collaborativescenario}

To test MR-HRC v2 and evaluate our hypotheses, we created a warehouse setting where humans share their working environment with a robot\footnote{\url{https://youtu.be/Ni-a2fXTb7o}}. 
The workspace is a room with three shelves and two crates, arranged to force humans and robots to cross paths unintentionally throughout the experiment, see Fig. \ref{fig::experiment}. 
The human teammate is supposed to restock \textit{Shelf 3} with bottles taken from \textit{Crate A} while the robot prepares an order by picking objects from \textit{Shelf 1} and \textit{Shelf 2} and placing them into \textit{Crate B}.

For the human task, twelve bottles with random numeric labels are inside \textit{Crate A}, shuffled before the beginning of each experiment.
The human operator should pick one item at a time from the crate and place it on \textit{Shelf 3}, rearranging the bottles by labels in descending order. 
Four bottles are distributed between \textit{Shelf 1} and \textit{2}, with one bottle intentionally placed out of the robot's reach. 
Bottle positions on the shelves are not predetermined, and the robot randomly chooses which bottle to reach and pick next.
Robot actions may fail (i.e. if an object slips from the gripper) or grasping it may be unfeasible (i.e. when the bottle is outside the robot's workspace).
In these cases, the human should understand the occurrence of the critical situation
and supervise the behaviour of the robotic teammate, possibly assisting it
by correcting its grasp or collecting the unreachable bottle and handing it over. 
While the robot is designed to be autonomous, the confined and intricate nature of the workspace can occasionally lead to situations where the robot and human unintentionally obstruct each other's movements, disrupting their tasks. To mitigate these issues, the human worker acts as a supervisor for the robot, monitoring its actions and intervening when necessary.
It is worth noting that, given the presence of one bottle outside the robot workspace, the human is forced to perform a handover with the robot once during the experiment. 
In these cases, the robot is instructed to reach a predefined handover pose (see Fig. \ref{fig:hrc-dt}) and wait for the human to bring the object.
The experiment is complete when the human has arranged all bottles on \textit{Shelf 3}, and the robot has filled \textit{Crate B} with the other four bottles. 

\begin{figure}[t!]
\centering 
\begin{subfigure}{.5\textwidth}
  \centering
    \includegraphics[width=0.65\textwidth]{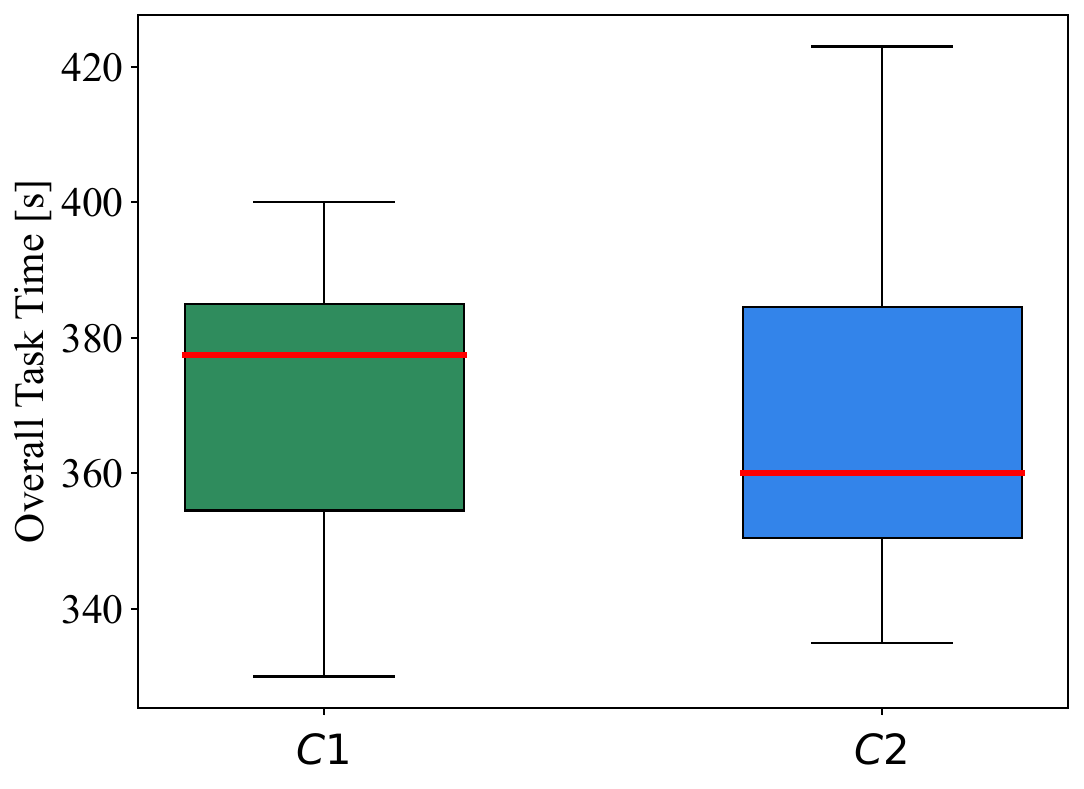}
    \caption{}
    \label{fig::overall-task-time}
\end{subfigure}
\begin{subfigure}{.5\textwidth}
  \centering
    \includegraphics[width=0.65\textwidth]{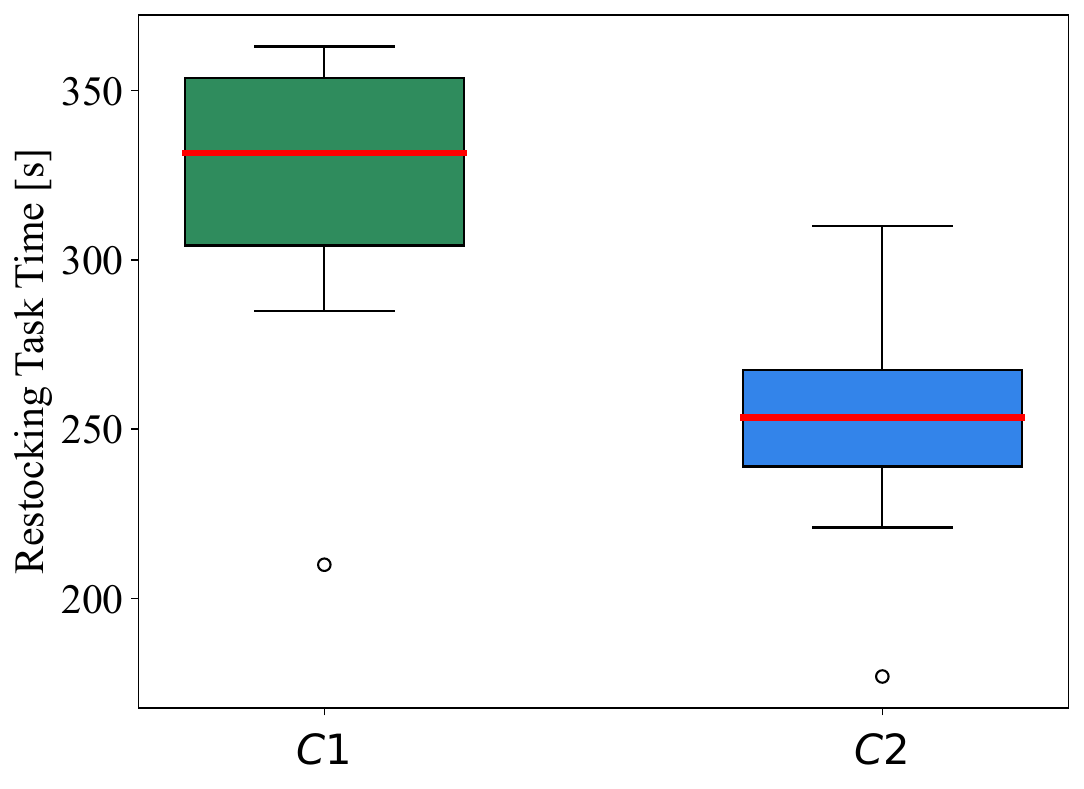}
    \caption{}
    \label{fig::human-task-time}
\end{subfigure}
\caption{ Time metrics (in seconds) observed during trials, in the two experimental conditions. Fig. \ref{fig::overall-task-time} depicts the total time needed to complete the collaboration, measured once the human and the robot had both completed their tasks. Conversely, Fig. \ref{fig::human-task-time} depicts the time taken by participants to complete their restocking task, measured once the human had put all twelve bottles on \textit{shelf 3} in the correct order. The small circles depicted in Fig. \ref{fig::human-task-time} highlight an outlier in the distribution.}
\label{fig:task-time}
\end{figure}

\begin{figure}[t!]
\centering
\begin{subfigure}{.35\textwidth}
\centering
\includegraphics[width=\linewidth]{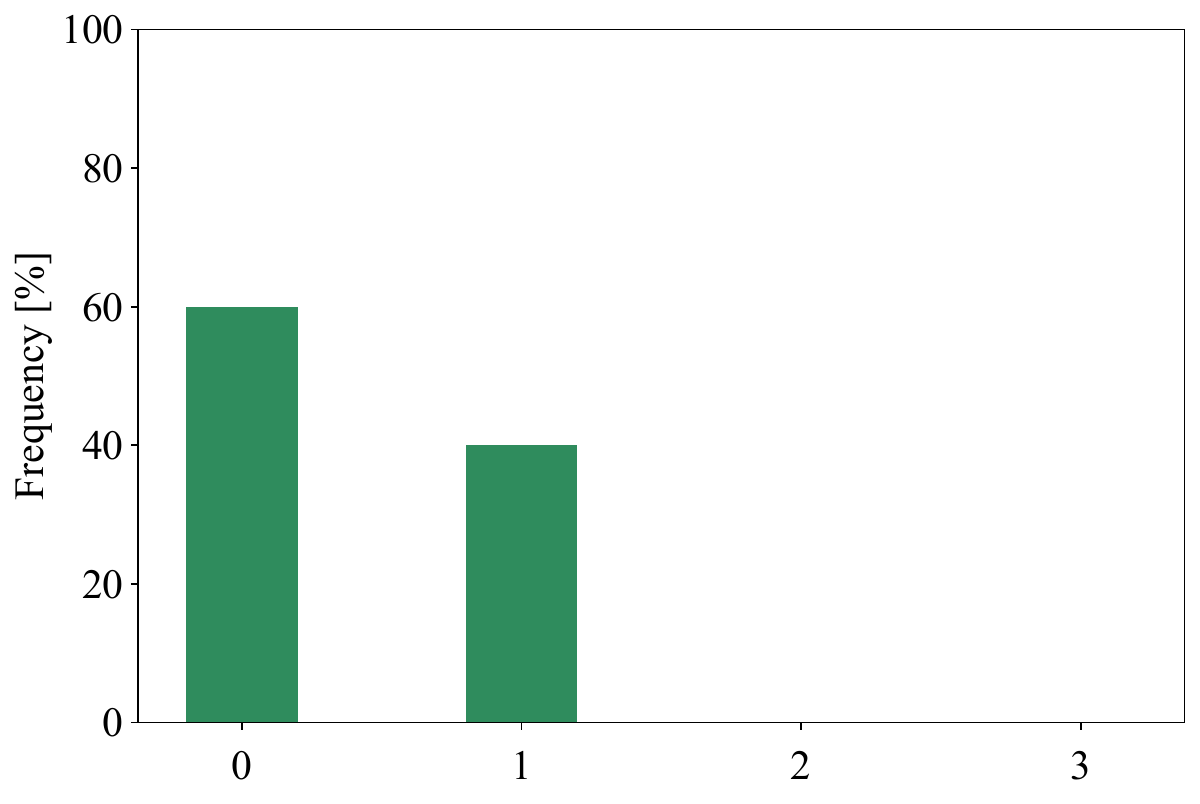}  
\caption{Human assistive interventions under \textit{C1}.}
\label{fig:assistance-no-mr}
\end{subfigure}
\begin{subfigure}{.35\textwidth}
\centering
\includegraphics[width=\linewidth]{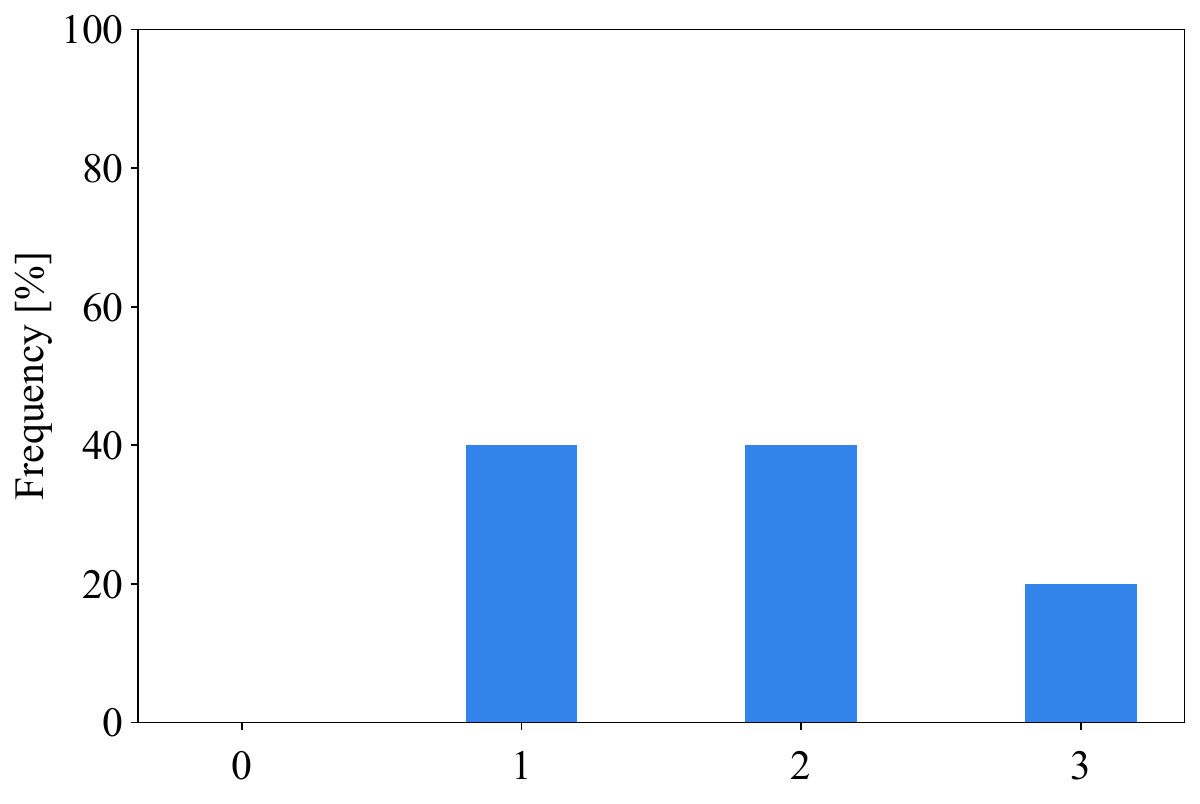}  
\caption{Human assistive interventions under \textit{C2}.}
\label{fig:assistance-mr}
\end{subfigure}
\caption{Histograms depicting the number of human proactive interventions per trial.}
\label{fig:assistance-no-mr-vs-mr}
\end{figure}

\subsection{User Study} 
\label{experiments}

We carried out an experimental campaign with $N=20$ participants (15 males and 5 females), all aged between 21-33 (\textit{Avg} $=23.7$, \textit{StdDev} $= 2.49$). 
All participants had little or no prior experience in terms of MR-HMDs and interaction with robotic platforms.
The experiment has been approved by the Ethical Committee for research at the University of Genoa through protocol n. 2021/65, issued on November 18, 2021.

Participants were asked to complete the joint activity with TIAGo++
(\textit{C1}) without MR communication, or 
(\textit{C2}) with MR communication.
Participants were randomly assigned to one of the two conditions.
Before the experiment, candidates were instructed on their tasks and how to interact with the robot. 
For participants performing under \textit{C2}, a very brief overview of how to navigate the holographic menus of the HMD was provided, along with instructions for the initial calibration of the MR application. 
After that, the experiment could start. 

Given the preliminary nature of the study and the complexity of the experimental scenario, we chose to focus our analysis exclusively on the evaluation of \textit{H1} and \textit{H2}, which only considers metrics associated with team fluency.
The two current hypotheses have been tested independently, using quantities measured during the experimental campaign. In particular, for
\textit{H1}, we manually timed each trial and measured how long it took for the human operator and the robot to complete their respective tasks. 
Each experiment was video recorded and subsequently analyzed offline by a coder to extract the following metrics related respectively to \textit{H2.1}, \textit{H2.2} and \textit{H2.3}:
\begin{itemize}
\item[\textit{M1}] Proactive human interventions - This metric measures the number of times a participant proactively assisted the robot during task execution. Examples include correcting a bottle's position for a more stable grasp or placement in the delivery box.
\item[\textit{M2}] Number of failed interactions - This metric captures instances where human-robot collaboration deviated from its intended course. Here is a breakdown of the categories that contribute to this metric: 
\begin{itemize}
\item[$\bullet$] Path Interference and Positioning Errors - This occurs when the human and robot physically block each other's movements (path interference). This disrupts the robot's navigation, causing it to deviate from its planned path and potentially leading to positioning errors (failing to reach the intended location for picking up or delivering a bottle).
\item[$\bullet$] Human Intervention Failure - These are situations where the robot is failing a task and the human does not intervene to correct it.
\end{itemize}
\item[\textit{M3}] Time Spent Monitoring the Robot - This metric measures the total occurrences when participants paused their task to evaluate the robot's behaviour. It reflects the participant's uncertainty about the robot's intention.
\end{itemize}


\section{Results}
\label{sec:results}


The results of the user study are hereby reported, with a focus on the two aforementioned hypotheses. 

As for \textit{H1}, Fig. \ref{fig::overall-task-time} and \ref{fig::human-task-time}, respectively, report the results related to completion time for the overall collaboration and completion time for the human restocking task only. 
It is possible to note in Fig. \ref{fig::overall-task-time} that the time required to complete the collaborative task remained comparable in both experimental conditions, typically ranging between 355 and 387 seconds. 
However, Fig. \ref{fig::human-task-time} shows that participants under \textit{C2} completed their restocking task, on average, in around 250 seconds, whereas the average measured time in condition \textit{C1} was around 330 seconds. While these numbers may appear quite large for such a simple restocking task, it is to be noted that participants were also simultaneously required to supervise the robot's actions and intervene when necessary.
\begin{figure}[t!]
\centering
\begin{subfigure}{.35\textwidth}
\centering
\includegraphics[width=\linewidth]{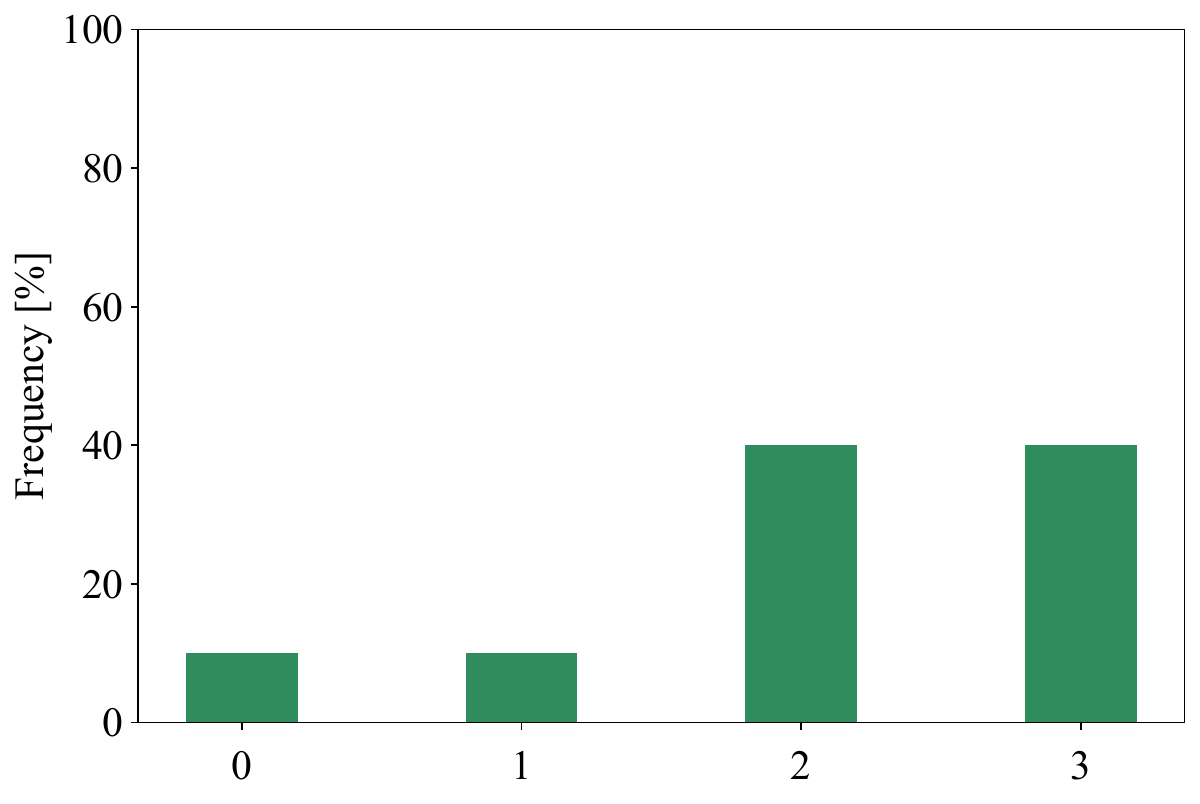}  
\caption{Failed interactions under \textit{C1}.}
\label{fig:failures-no-mr}
\end{subfigure}
\begin{subfigure}{.35\textwidth}
\centering
\includegraphics[width=\linewidth]{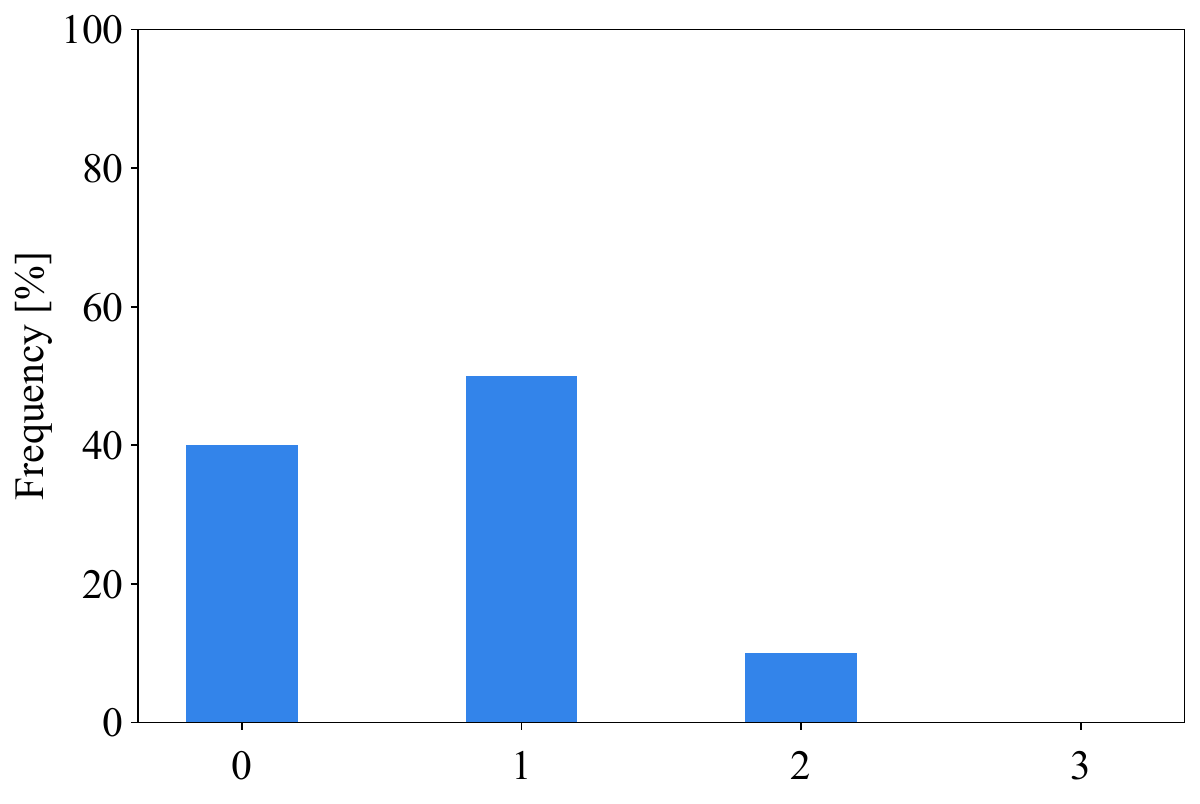}  
\caption{Failed interactions under \textit{C2}.}
\label{fig:failures-mr}
\end{subfigure}
\caption{Histograms depicting the number of failed interactions per trial.}
\label{fig:failures-no-mr-vs-mr}
\end{figure}

As such, an interpretation of the results conveyed by Fig. \ref{fig::overall-task-time} and \ref{fig::human-task-time} is given as follows:
participants always completed their tasks before the robot, and the total completion time depended only on the robot's performance, which was comparable in the two conditions. 
Nevertheless, participants in condition \textit{C2}, aware of the robot's upcoming intentions thanks to the holographic interface, managed to plan their movements and actions synchronously with those of their robot teammate, resulting in fewer mutual obstructions and faster completion times on the human's side. 
To further validate such results, we performed a t-test on the distributions depicted in Fig. \ref{fig::overall-task-time} and \ref{fig::human-task-time}, which could be assumed normal through the Shapiro-Wilk test \cite{shapiro1965analysis} ($p$-value $> 0.05$ for all distributions). 
For the data in Fig. \ref{fig::overall-task-time}, the t-test returned $p$-value $> 0.4$, confirming that no significant difference could be observed in the total completion time in conditions \textit{C1} or \textit{C2}. 
Conversely, the t-test performed on the distributions in Fig. \ref{fig::human-task-time} yielded $p$-value $< 0.01$, thus corroborating the significant difference between times measured on completion of the restocking task under \textit{C1} or \textit{C2}.

To evaluate hypothesis \textit{H2}, we adopted the various metrics illustrated in the previous Section and compared them under the two experimental conditions. 
Fig. \ref{fig:assistance-no-mr-vs-mr} reports the results related to the amount of human assistance offered to the robot (M1) \color{black}. 
The histograms specifically show the distribution of participants across different numbers of proactive interventions to assist the robot in completing the collaboration task. 
For example, in condition \textit{C1}, $60\%$ of participants concluded their experiment without helping the robot. The remaining $40\%$ only intervened once. 
In contrast, participants in condition C2, who received anticipatory holographic communication, exhibited greater proactiveness. This is reflected in the distribution: $40\%$ intervened at least once, another $40\%$ intervened twice, and the remaining participants intervened three times.
Since data distribution in Fig. \ref{fig:assistance-no-mr-vs-mr} could not be assumed normal, we adopted the non-parametric Wilcoxon signed-rank test \cite{wilcoxon1992individual}.
The test yielded a statistic $W = 36$, which should be compared with the critical value $W_c = 60$ extracted from \cite{wilcoxon1947probability}, fixing the significance level $\alpha = 0.05$ and the sample size $N=20$. 
Since $W < W_c$, we could reject the null hypothesis and confirm the significant difference between the degree of human assistance in conditions \textit{C1} and \textit{C2}. It is important to note that the software and hardware controlling the robot's behaviour were identical in both conditions (C1 and C2). This ensures that the observed differences in human intervention rates stem from the communication method (anticipatory holographic vs. no) and not from potential performance issues with the robot in C2.

\begin{figure}[t!]
\centering
\includegraphics[width=0.32\textwidth]{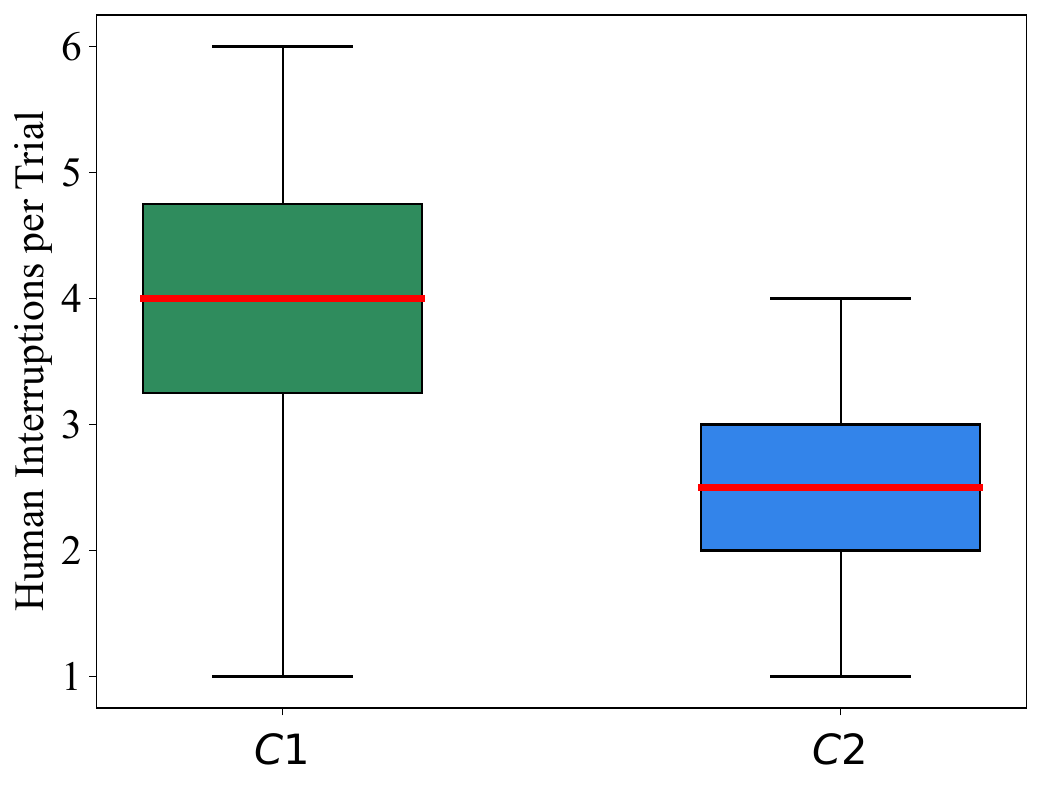}
\caption{
Number of times per trial in which participants paused their task to observe the robot.}
\label{fig::interruptions}
\end{figure}

Similar considerations can be made by observing Fig. \ref{fig:failures-no-mr-vs-mr}, which depicts data related to failed interactions during the experiments (M2). 
Under condition \textit{C1}, participants lacked an intuitive communication channel with the robot. This resulted in a higher number of failed interactions: $40\%$ failed three times, $40\%$ failed twice, $10\%$ failed once, and only $10\%$ avoided any failure.
In contrast, under condition \textit{C2} participants were more aware and responsive due to the improved communication method. This led to a significant decrease in failed interactions: $40\%$ had no failure, $50\%$ failed only once, and the remaining $10\%$ failed twice.
Again, we employed the Wilcoxon test to evaluate the significance of such results. 
In this case, the test yielded statistic $W = 4$, and by comparing this value with the critical one ($W_c$) mentioned before, we could confirm the statistical difference between the two experimental conditions.

Fig. \ref{fig::interruptions} illustrates the number of times participants interrupted their restocking task to observe the robot (M3). 
 In condition C1, participants interrupted their restocking task an average of four times per trial. This frequent interruption suggests that participants found it difficult to infer the robot's intention without a holographic communication channel. In contrast, when the holographic interface is introduced (condition \textit{C2}) interruptions significantly reduce, $75\%$ of participants paused their task less than three times. This aligns with the shorter task time observed in Fig. \ref{fig::human-task-time}.
 We assessed the significance of these results through a t-test, carried out after ensuring that distributions could be assumed normal (Shapiro-Wilk test yielded $p$-values $> 0.2$ for both cases). 
The t-test returned a $p$-value $< 0.01$, corroborating the statistical difference between \textit{C1} and \textit{C2}.


\section{Conclusions}
\label{sec:conclusions}

We presented MR-HRC v2, a software architecture that utilizes Mixed Reality to facilitate smooth and intuitive collaboration between humans and robots. MR serves as a communication layer that intuitively conveys the robot's intentions and forthcoming actions to the human collaborator. This approach, which builds on anticipatory communication, was originally proposed by Macciò \textit{et al}., 2022 \cite{maccio2022mixed}, and has been expanded to include scenarios with mobile manipulators, enabling the conveyance of navigation intentions through moving holograms. The architecture was evaluated in a user study involving human interaction with TIAGo++, a state-of-the-art dual-arm manipulator from PAL Robotics, within a collaborative environment. Such an experimental campaign allowed us to evaluate the effectiveness of MR as a medium to communicate the robot's intentions, comparing two experimental conditions. In particular, we found that participants experiencing the whole holographic communication (condition \textit{C2}) accomplished a smoother collaboration with the robot. This resulted in fewer mutual hindrances and improved awareness of the teammate's actions, which may lead to safer collaborative conditions.

Future studies in this direction should also assess the subjective perception of humans regarding the interaction with the robot and the holographic communication interface.

\section*{Acknowledgement}
This research was partially supported by the Italian government under the National Recovery and Resilience Plan (NRRP), Mission 4, Component 2 Investment 1.5, funded from the European Union NextGenerationEU and awarded by the Italian Ministry of University and Research. Furthermore, it has been also partially supported by the NVIDIA Academic Hardware Grant Program.




\bibliographystyle{IEEEtran}
\bibliography{bibliography}

\end{document}